\documentclass{article}
\usepackage{ijcai24}

\usepackage{times}
\usepackage{soul}
\usepackage{url}
\usepackage[hidelinks]{hyperref}
\usepackage[utf8]{inputenc}
\usepackage[small]{caption}
\usepackage{graphicx}
\usepackage{amsmath}
\usepackage{amsthm}
\usepackage{booktabs}
\usepackage{algorithm}
\usepackage{algorithmic}
\usepackage[switch]{lineno}
\usepackage{xspace}
\usepackage{amssymb}
\usepackage{graphicx}
\usepackage{multirow}
\usepackage{xcolor}
\usepackage{pgfplots}
\usepackage{float}
\usepackage{subfigure}
\usepackage{color}

\newcommand{\model}{LPDGAN\xspace}
\newcommand{\dataset}{LPBlur\xspace}

\title{A Dataset and Model for Realistic License Plate Deblurring}

\author{
Haoyan Gong
\and
Yuzheng Feng\and
Zhenrong Zhang\and
Xianxu Hou\and
Jingxin Liu\and
Siqi Huang\And
Hongbin Liu \thanks{Corresponding author}\\
\affiliations
School of AI and Advanced Computing, Xi'an Jiaotong-Liverpool University\\
\emails
\{\href{mailto:haoyan.gong21@student.xjtlu.edu.cn}{haoyan.gong21}, \href{mailto:yuzheng.feng21@student.xjtlu.edu.cn}{yuzheng.feng21}, \href{mailto:zhenrong.zhang21@student.xjtlu.edu.cn}{zhenrong.zhang21}\}@student.xjtlu.edu.cn,
\{\href{mailto:xianxu.hou@xjtlu.edu.cn}{xianxu.hou}, \href{mailto:jingxin.liu@xjtlu.edu.cn}{jingxin.liu}, \href{mailto:siqi.huang@xjtlu.edu.cn}{siqi.huang}, \href{mailto:hongbin.liu@xjtlu.edu.cn}{hongbin.liu}\}@xjtlu.edu.cn
}

\begin{document}

\maketitle

\begin{abstract}
    Vehicle license plate recognition is a crucial task in intelligent traffic management systems. However, the challenge of achieving accurate recognition persists due to motion blur from fast-moving vehicles. Despite the widespread use of image synthesis approaches in existing deblurring and recognition algorithms, their effectiveness in real-world scenarios remains unproven. To address this, we introduce the first large-scale license plate deblurring dataset named License Plate Blur (\dataset), captured by a dual-camera system and processed through a post-processing pipeline to avoid misalignment issues. Then, we propose a License Plate Deblurring Generative Adversarial Network (\model) to tackle the license plate deblurring: 1) a Feature Fusion Module to integrate multi-scale latent codes; 2) a Text Reconstruction Module to restore structure through textual modality; 3) a Partition Discriminator Module to enhance the model's perception of details in each letter. Extensive experiments validate the reliability of the \dataset dataset for both model training and testing, showcasing that our proposed model outperforms other state-of-the-art motion deblurring methods in realistic license plate deblurring scenarios. The dataset and code are available at \href{https://github.com/haoyGONG/LPDGAN}{\textit{https://github.com/haoyGONG/LPDGAN}}.

\end{abstract}

\section{Introduction}

Efficient recognition of vehicle license plates is crucial for intelligent traffic management systems, however real-world scenarios often pose a significant challenge due to motion blur. This blur, making license plates unreadable, is especially problematic in situations involving high-speed vehicles or low-light conditions. Such issues are exacerbated during nighttime or in bad weather, resulting in considerable motion blur in captured images. To tackle these issues, our study introduces a comprehensive dataset and a novel model tailored for realistic license plate deblurring. 

Image deblurring is a key task in computer vision, focused on restoring blurred images to clear ones for accurate observation and identification. The progress in this field heavily depends on the development of relevant datasets. Current methods for creating deblurring datasets fall into three main categories: (1) synthetic blurring using blur kernels ~\cite{sun2013edge,kohler2012recording,lai2016comparative}, which leads to a lack of generalization capability for models trained on these synthesized images when applied to real-world images. (2) The generation of blurred images from sharp frames via averaging or fusion~\cite{nah2017deep,shen2019human,jiang2020learning}, which doesn’t fully mimic real-world overexposure outliers ~\cite{chang2021beyond}. (3) Lastly, beam-splitting systems capture sharp and blurred image pairs via camera shake~\cite{rim2020real}, with potential issues in color accuracy and alignment. Each approach contributes to the field but also has inherent limitations impacting the realism and utility of the datasets.

\begin{figure}[t]
		\centering
		\includegraphics[width=1\linewidth]{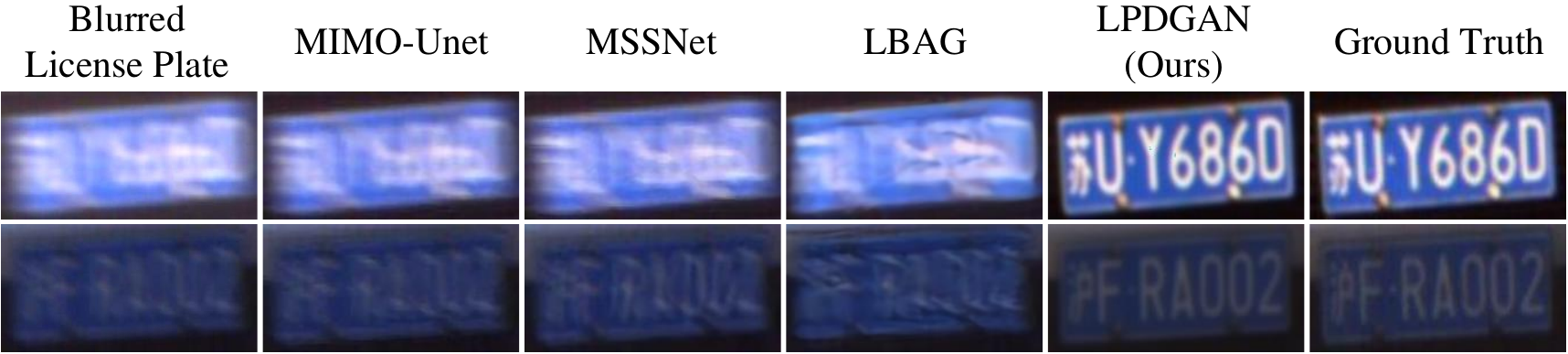} 
	\caption{The visual deblurring results of several state-of-the-art models and our model for real-world motion blurred license plate images.}
	\label{Fig:pretrain}
\end{figure}

With the advent of deep learning, numerous convolutional neural network (CNN)-based methods have surfaced \cite{sun2015learning,gong2017motion,tao2018scale,shen2019human,zhang2023generalizing}, playing an essential role in the motion deblurring task. Recently, the proposal of Generative Adversarial Networks (GAN) has also profoundly impacted the image deblurring field \cite{ramakrishnan2017deep,kupyn2018deblurgan,kupyn2019deblurgan,zhao2022fcl}. Despite these advancements, deblurring license plate images remains a significant challenge, primarily due to the lack of large-scale, tailored datasets. The complexity of license plate blurring, with its more severe degradation compared to standard motion blur, poses an additional challenge. To better justify the performance of existing image deblurring algorithms on real-world blurred license plate images, we evaluate the performance of several state-of-the-art deblurring algorithms with blurred license plate images. As shown in Figure \ref{Fig:pretrain}, we can conclude that all these methods fail to perform well in this task. 
It underscores the necessity for further research specifically targeting real-world vehicle license plate deblurring.

To address these challenges, we present a comprehensive solution consisting of a large-scale paired license plate dataset and a dedicated license plate deblurring model. Our data collection employs a dual-camera setup with different shutter speeds to capture sharp and blurred images simultaneously, eliminating color deviations and enabling post-processing alignment. Our innovative end-to-end model leverages an encoder and latent fusion module for handling multi-scale latent codes, featuring the Swin transformer block (\cite{liu2021swin}) for effective long-range modeling. To enhance letter reconstruction and text legibility, we introduce a partition discriminator assessing per-letter sharpness. Extensive experiments using our \dataset dataset, including metrics such as $L_1$ loss, Peak Signal-to-Noise Ratio (PSNR), Structural Similarity Index (SSIM), Perceptual Loss (PerL), and Text Levenshtein Distance (TLD) \cite{levenshtein1966binary}, validate its suitability for training and testing. Our proposed model outperforms state-of-the-art motion deblurring methods in real-world license plate deblurring scenarios.

In summary, our main contributions are as follows:
\begin{itemize}
    \item  We present a real-world sharp-blurred license plate dataset, named \dataset. This dataset consists of 10,288 paired images, meticulously collected under diverse real-road scenarios using our designed dual-camera system, and corrected by a post-processing pipeline.
    \item We introduce a novel \model, a license plate deblurring model that leverages multi-scale latent codes as references. It incorporates both a partition discriminator and text reconstruction techniques, which enhance the model's capability to generate high-quality license plate images through spatial architecture and textual information, respectively.
    \item Extensive experiments demonstrate that our dataset \dataset is highly effective for model training and evaluation. Compared to other state-of-the-art (SOTA) deblurring models, our proposed \model can achieve 21.24\% license plate recognition accuracy improvement. 
\end{itemize}
\begin{figure*}[t]
		\centering
		\includegraphics[width=1\linewidth]{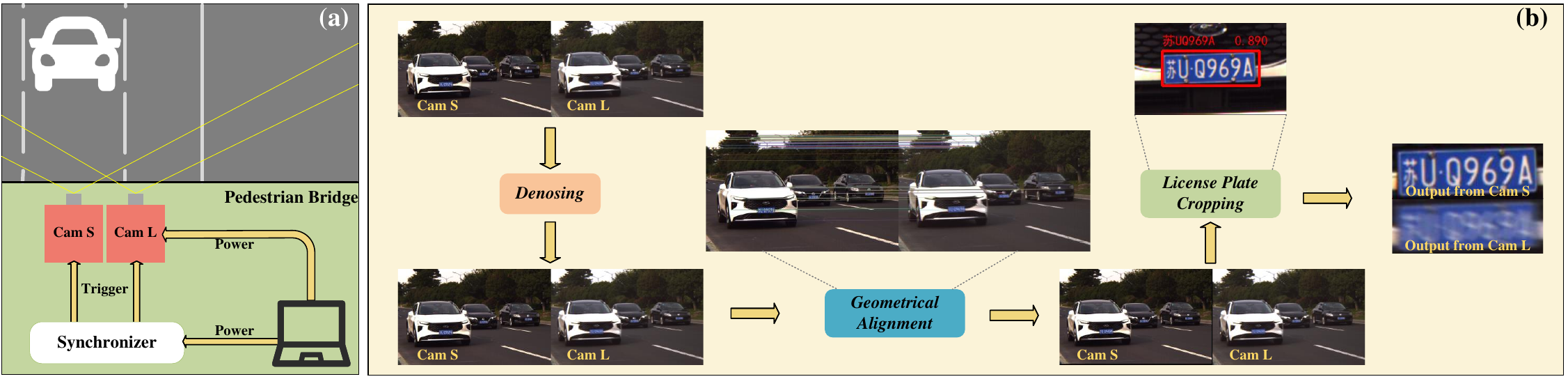} 
	\caption{(a) A schematic diagram of the paired image acquisition system collecting data in a pedestrian bridge. (b) The pipeline of paired images post-processing.}
	\label{Fig:data-collection}
\end{figure*}

\section{Related Work}
\label{section: Related Work}
\subsection{Image Deblurring Datasets}
Image deblurring relies on paired sharp-blur image datasets. Traditionally, blurred images are generated by convolving sharp images with uniform or non-uniform blur kernels \cite{levin2009understanding,sun2013edge,kohler2012recording,lai2016comparative}. Consequently, some researchers attempt to capture sequences of sharp frames while vibrating the camera, averaging or fusing such sequences of frames into corresponding motion-blurred images \cite{nah2017deep,shen2019human,jiang2020learning,noroozi2017motion}. HIDE dataset \cite{shen2019human} is created by averaging 11 consecutive frames, with the central frame serving as the sharp image. The same strategy is employed in the collection of the Blur-DVS dataset \cite{jiang2020learning} and MSCNN (WILD) dataset \cite{noroozi2017motion}. However, models lack generalizability to real-world scenarios when they are trained on such synthetic datasets generated using the aforementioned methods. Recently, certain researchers gathered authentic pairs of sharp-blur images employing beam-splitting systems. They position two cameras at a fixed angle to ensure that both images share the same visual field, as described in works by \citeauthor{rim2020real} and \citeauthor{li2023real}.
However, this approach can lead to color cast discrepancies in the paired images due to inherent issues with beam-splitting systems.

\subsection{Blind Deblurring}
The majority of conventional approaches employ priors of natural images to estimate latent images or blur kernels \cite{fergus2006removing,shan2008high,cho2009fast,ren2018deep,whyte2012non}. However, the aforementioned techniques have certain limitations by predicating upon the assumption of uniform image blur. To address this issue, some methods \cite{ren2017partial,hyun2013dynamic} estimate blur kernels at a pixel level, thereby accommodating more complex blurring situations.

With the advent of deep learning technologies, significant strides have been made in image deblurring, applying deep learning to predict blur kernels or latent images to procure clear images. In the work of \cite{sun2015learning}, a method based on CNNs is proposed to predict the probability distribution of block-level motion blur. \citeauthor{gong2017motion} introduces a method to directly estimate the motion flow of blurred images, recovering non-blurred images from the estimated motion flow.
MIMO-UNet \cite{cho2021rethinking} deploys a multi-input-multi-output single Unet network to simulate multi-level Unet for noise reduction across various image scales. MSSNet \cite{kim2022mssnet} enhances deblurring network performance by using a stage configuration reflecting blur scales, an inter-scale information propagation scheme, and a pixel-shuffle-based multi-scale scheme. XYDeblur \cite{ji2022xydeblur} further augments network efficiency and deblurring performance by employing rotated and shared kernels within the decoder.

\subsection{GAN-Based Deblurring}
In recent years, following the inception of GANs, their application in the domain of image deblurring has achieved remarkable success. DeblurGAN \cite{kupyn2018deblurgan} first presents an end-to-end learning method for motion deblurring, and also introduces a new method for blur generation. DeblurGAN-v2 \cite{kupyn2019deblurgan} introduces a dual-scale discriminator based on a relative conditional GAN framework and incorporates a feature pyramid into the deblurring process, which permits the flexible substitution of the backbone network.
MSG-GAN \cite{karnewar2020msg} addresses the issue of insufficient overlap between the true and false support distributions during the transfer from discriminator to generator in GANs by allowing multi-scale gradient networks from the discriminator to the generator. FCL-GAN \cite{zhao2022fcl} designs a lightweight domain conversion unit (LDCU) and a parameter-free frequency-domain contrastive unit (PFCU) for lightweight property and performance superiority. The aforementioned methods can handle standard blurred images, but they struggle to deliver satisfactory results on license plate blurring with very severe degradation. We propose an end-to-end generative model that accommodates multi-scale inputs and outputs, employing several novel modules to accomplish the license plate deblurring task better. 

\begin{figure*}[t]
		\centering
		\includegraphics[width=1\linewidth]{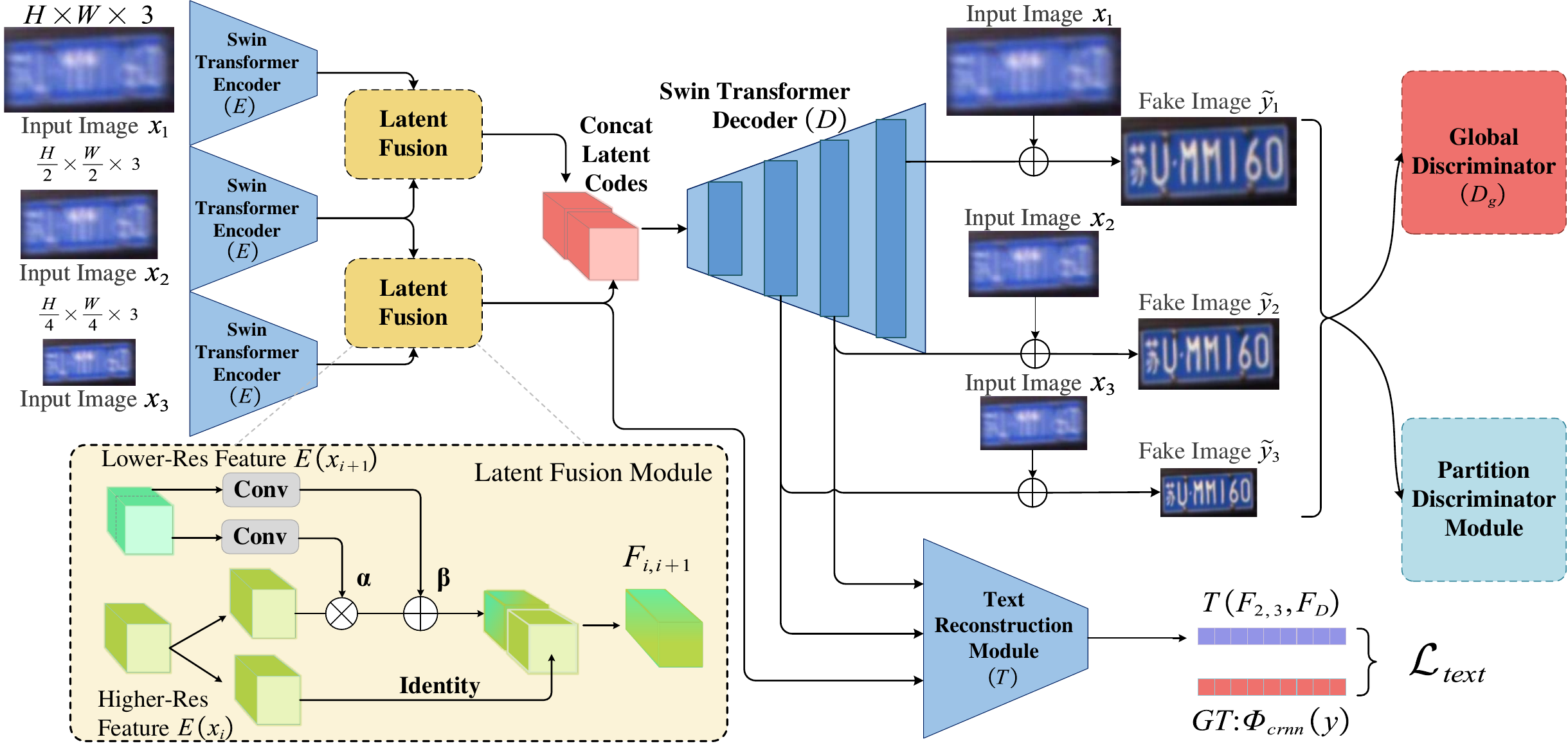} 
	\caption{Overview of the proposed Licence Plate Deblurring Generative Adversarial Network.}
	\label{Fig:overview model}
\end{figure*}

\section{Proposed \dataset Dataset}
\subsection{Data Collection}

\paragraph{Causes of Motion Blur.} 
Motion blur refers to the perceptible streaking effect observed when capturing the movement of objects.
In the capturing process, the correlation between the amount of light entering the photosensitive component and the camera's basic parameters satisfies the following equation:
\begin{equation}
\small
La\,\,\propto \,\,S_L\times ISO\times Et\times \left( Ap \right) ^2,
\label{equ:light amount}
\end{equation}
where ${La}$ denotes the number of photons received by the camera, ${S_L}$ represents the light intensity of the scene, ${ISO}$ represents the camera ISO value, ${Et}$ is the exposure time, ${Ap}$ denotes the camera aperture size. Cameras adjust these parameters automatically within limits depending on the lighting situation. For example, cameras increase their aperture and exposure time in low-lighting settings to capture enough light. Fast-moving objects leave trajectories within a single frame during this extended exposure time, resulting in a motion blur effect.  

\paragraph{Paired Image Acquisition.} 
To collect paired sharp-blur images, we use two identical scientific cameras that are set with different exposure times. As shown in Figure \ref{Fig:data-collection}a, cameras S and L are fixed parallel on a tripod to maintain horizontally to the ground. Specifically, Camera S is set by an extremely short exposure time ${Et}_{s}$, employed for the collection of sharp images, while Camera L is set by a relatively longer exposure time ${Et}_{l}$ for the acquisition of blurred images. Both of these cameras are interfaced with a computer via a synchronizer, which ensures the synchronization of the start of exposures, and both cameras capture the same scene.

Scenes are taken in a variety of locations, including above, on the right side, and on the left side of roadways, to guarantee the dataset's diversity. Also, depending on the road and the illumination conditions, we dynamically modulate camera exposure time according to the subsequent equation:

\begin{equation}
\small
v\cdot Et=\frac{b\cdot D}{p\cdot f},
\label{equ:blur distance}
\end{equation}
where ${v}$ denotes vehicular velocity, ${Et}$ represents exposure time, ${b}$ represents pixels blurred, ${D}$ is the distance between vehicle and camera, ${f}$ denotes camera focal length, and ${p}$ is pixel edge length on the sensor. Given that the actual speeds of individual vehicles are indeterminable, we standardize image captures on high-speed road sections with a regulatory speed limit of 70 km/h to ensure that ${D}$ is remained within a range of 10-20 meters.

Moreover, to ensure equality in exposure between two cameras, we make their exposure times and ISO values to satisfy the following equation:
\begin{equation}
\small
\frac{ISO_s}{ISO_l}=\frac{Et_l}{Et_s},
\label{equ:ISO time}
\end{equation}
where ${ISO}_{s}$ is the ISO value for Camera S and ${ISO}_{l}$ Camera L. However, variations in ISO values can cause changes in image noise levels, in post-processing, we incorporate a denoising step for sharp images.

\subsection{Data Post-processing}
As shown in Figure \ref{Fig:data-collection}b, the paired image post-processing includes denoising, geometrical alignment, and license plate cropping.
\paragraph{Denoising.} 
Due to the disparate ISO settings of the two cameras, Camera S and Camera L capture images with a different noise level. Consequently, during the conversion of RAW images to RGB format, wavelet denoising \cite{liu2020densely} is employed after white-balancing, color mapping, and gamma correction. 
\paragraph{Geometrical Alignment.} 
Cameras S and L capture sharp and blurred image pairs with slight horizontal misalignment even though they are closely aligned left-to-right to minimize the difference. To align these image pairs perfectly, we first take a static image pair without any moving vehicles as the reference pair for each scene. Then, the Enhanced Correlation Coefficient Maximization \cite{evangelidis2008parametric} is adopted to estimate the geometric transformation between the sharp and blur of the reference image pair. Finally, the estimated geometric transformation is applied to the image pairs in the same scene.  

\paragraph{License Plate Cropping.} 
A pre-trained YOLO v5 \cite{Jocher_YOLOv5_by_Ultralytics_2020} and a pre-trained CRNN \cite{shi2016end} model are facilitated the detection and recognition of license plates under standard conditions, both models are pre-trained on the CCPD \cite{xu2018towards} license plate dataset. Following the geometrical alignment, the pre-trained YOLO v5 and CRNN detect and recognize the bounding box of each sharp plate in the paired images, both the sharp and blurred images are then cropped using the same detected coordinates.

In conclusion, we collect 10,288 image pairs, with an original image size of $1920\times 1220$. Post-processing crops image size to $224\times 112$ with blur size ranging from 20-50 pixels. Among them, 5672 pairs are captured under normal light conditions and 4616 pairs under low light conditions, including 1,000 pairs under rainy environmental conditions. For more information, please refer to the released dataset on the GitHub repository.

\section{Method}
\paragraph{Overview.}
The goal of our work is to improve the clarity of license plate images using a meticulously crafted image-to-image translation framework, called \model. As depicted in Figure \ref{Fig:overview model}, our approach first constructs a multi-scale feature extraction and fusion module designed to encode input blurred images effectively. Subsequently, an image decoder is employed to generate sharp and high-quality images. To further enhance the overall image quality, we integrate both a global discriminator and a partition discriminator for adversarial training. Additionally, we incorporate a text reconstruction module to enrich the semantic information embedded in the generated license plate images. 

\subsection{Multi-scale Feature Extraction and Latent Fusion Module}
\paragraph{Feature Extraction.}
In real-world scenarios, license plate images affected by motion blur often exhibit intricate degradations, including noise, low resolution, and ghosting effects. Our feature encoder, denoted as $E$, is specifically used to address these degradations, extracting essential image features for subsequent processing. In particular, the Swin transformer block \cite{liu2021swin} is selected for its ability to capture global information through self-attention mechanisms. This is crucial to resolve the elongated ghosting artifacts that often appear in motion-blurred license plate images. To address variations in license plate image sizes due to differing capture distances, our approach employs a multi-scale feature extraction strategy, which is illustrated in Figure \ref{Fig:overview model}. This approach facilitates the encoding of features at each scale, which are represented as $E(x_i)$ for $i=1,2,3$.

\paragraph{Latent Fusion.}
Based on the Spatial Feature Transform (SFT) \cite{wang2018recovering}, we further propose a Latent Fusion Module $F$ (see Figure \ref{Fig:overview model}). This module is designed based on an affine transformation to effectively integrate the obtained multi-scale features. Specifically, for the fusion of $E(x_1)$ and $E(x_2)$, we first split $E(x_2)$ along the channel dimension. Each part is then processed through a series of convolutional layers to derive the fusion parameters $ \alpha $ and $ \beta $. These parameters are employed to modulate $E(x_1)$ through scaling and shifting operations. Moreover, we reintegrate the original $E(x_2)$ using a skip connection, which is then combined with the modified $E(x_1)$ along the channel dimension. This fusion process is also applied between $E(x_2)$ and $E(x_3)$. The corresponding formulas are provided below:
\begin{equation}
\begin{split}
\small
&\alpha ,\beta =\text{Conv}\left( E(x_{i+1})_{sp1}+E(x_{i+1})_{sp2} \right), \\
&F_{i,i+1}=\text{Concat}\left( \alpha \odot E(x_i)+\beta ,E(x_i) \right),
\end{split}
\label{equ:fusion}
\end{equation}
where $E(x_{i+1})_{sp1}$ and $E(x_{i+1})_{sp2}$ represent the two parts into which $E(x_{i+1})$ is divided along the channel dimension.
\subsection{Decoder and Discriminator}
\paragraph{Decoder.}
As depicted in Figure \ref{Fig:overview model}, our decoder $D$ is composed of a sequence of Swin transformer blocks and patch expanding operations. Similar to the encoder, our decoder is designed to generate sharp images in a multi-scale fashion, and the output images are denoted as $\tilde{y}_1$, $\tilde{y}_2$ and $\tilde{y}_3$ accordingly.

\begin{figure}[t]
		\centering
		\includegraphics[width=1\linewidth]{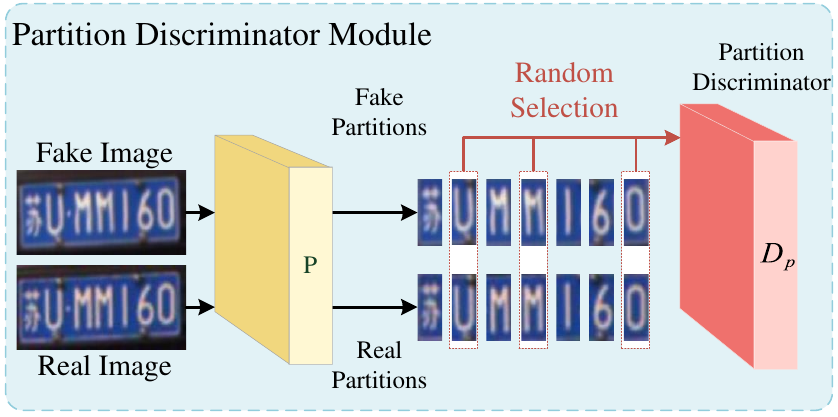} 
	\caption{The architecture of Partition Discriminator Module.}
	\label{Fig:partition}
\end{figure}

\begin{figure*}[t]
		\centering
		\includegraphics[width=1\linewidth]{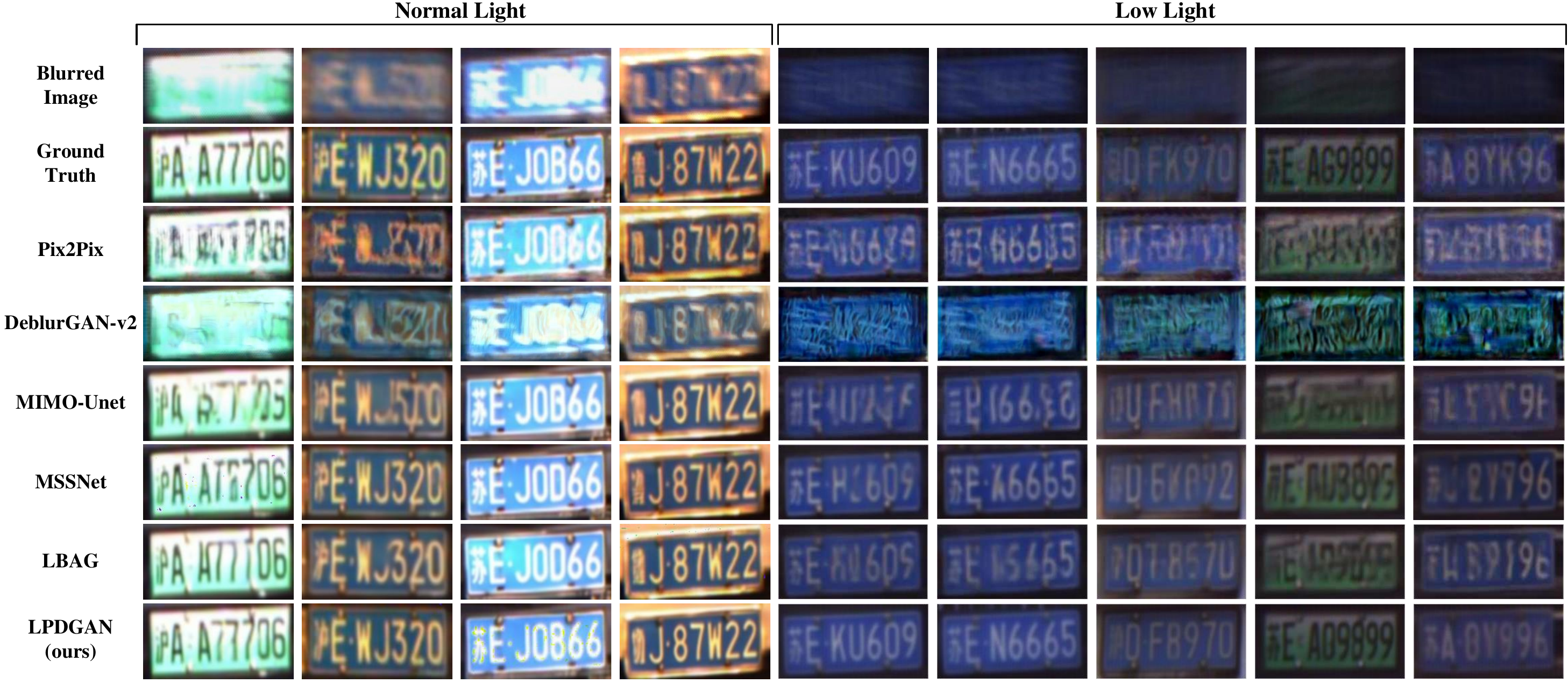} 
	\caption{Visual comparison of different deblurring methods on \dataset dataset. The test scene is divided into normal light and low light. Since the results in low light scenes are difficult to distinguish visually, we uniformly increase their brightness. The original brightness of the low light scene refers to the first line, which is Blur Image.}              
	\label{Fig:result}
\end{figure*}

\begin{table*}
\tabcolsep=0.023\linewidth
\caption{Quantitative results of comparing motion deblurring models. ``PerL" and ``TLD" denote Perceptual Loss and Text Levenshtein Distance, respectively.}
\begin{tabular*}{\linewidth}{ccccccccc}
   \toprule
   \multirow{2}*{Scenario} & \multicolumn{4}{c}{Normal Light} & \multicolumn{4}{c}{Low Light} \\
   \cmidrule(lr){2-5}\cmidrule(lr){6-9} & PerL$\downarrow$ & PSNR$\uparrow$ & SSIM$\uparrow$ & TLD$\downarrow$ & PerL$\downarrow$ & PSNR$\uparrow$ & SSIM$\uparrow$ & TLD$\downarrow$ \\
   \midrule
   Pix2Pix & 5.57 & 28.89 & 0.6669 & 1.35 & 2.24 & 28.71 & 0.7491 & 2.53\\
   DeblurGAN v2  & 8.02 & 28.51 & 0.5257 & 2.28 & 4.32 & 28.11 & 0.5451 & 4.34\\
   MIMO-UNet  & 3.79 & 29.12 & 0.7448 & 1.02 & 1.65 & 29.03 & 0.8083 & 2.68\\
   MSSNet  & 3.39 & 29.63 & 0.7891 & 0.62 & 2.74 & 29.62 & 0.8725 & 1.05\\
   LBAG  & 3.34 & 29.24 & 0.7916 & 0.58 & 1.44 & 29.44 & 0.8889 & 1.13\\
   \model(ours) & \textbf{3.31} & \textbf{29.95} & \textbf{0.7950} & \textbf{0.57} & \textbf{1.01} & \textbf{30.96} & \textbf{0.9214} & \textbf{0.81}\\
   \bottomrule
\end{tabular*}
\label{tab: total compare}
\end{table*}

\paragraph{Discriminator.}
 We design two discriminators: 1) the Global Discriminator ($D_{g}$) enhances overall spatial and color information in the restored images; 2) the Partition Discriminator ($D_{p}$) focuses specifically on refining character information by examining $n$ randomly selected partitions of letters within the license plate image. The structure of $D_{p}$ is shown in Figure \ref{Fig:partition}. It identifies and locates letter positions in both the real image $y$ and the fake image $\tilde{y}$. Following this, $n$ partition images are randomly chosen for evaluation by Partition Discriminator. In the early stages of training, when our mode's capacity to produce sharp images is still developing, the generated image might not be recognized with high accuracy. To address this, an average partitioning approach is applied to both $y$ and $\tilde{y}$, initially setting $n$ to $7$. As training progresses, a pre-trained YOLO v5 model is used for precise letter detection and the number of partitions n is set to 3. In our experiment, we employ WGAN-GP to train our model. In particular, the adversarial loss for the Global Discriminator can be formulated as follows:
\begin{equation}
\begin{split}
\mathcal{L}_{D_g}=&\mathbb{E}_{\tilde{y}}\left[D_{g}\left( \tilde{y} \right)\right]-\mathbb{E}_{{y}}\left[D_{g}\left({y} \right)\right] \\
&+\lambda_{gp1}\mathbb{E}_{\hat{y}}{\left[\left(\left\| \nabla_{\hat{y}} {D_{g}}\left( \hat{y} \right) \right\|_{2}-1\right)^2\right]}, \\
\hat{y}=\epsilon & \cdot  \tilde{y}+ \left(1-\epsilon  \right) \cdot y, \epsilon \sim U \left[0,1\right].
\end{split}
\label{equ:loss of Dg}
\end{equation}
Similarly, for the Partition Discriminator  $D_{p}$, the formulation is as follows: 
\begin{equation}
\begin{split}
\mathcal{L}_{D_p}=&\mathbb{E}_{\tilde{y}}\left[D_{p}\left( P\left(\tilde{y}\right) \right)\right]-\mathbb{E}_{{y}}\left[D_{p}\left(P\left( {y} \right) \right)\right]\\
&+\lambda_{gp2}\mathbb{E}_{\hat{y}}{\left[\left(\left\| \nabla_{\hat{y}} {D_{p}}\left( P\left( \hat{y} \right) \right) \right\|_{2}-1\right)^2\right]},\\
\hat{y}=\epsilon & \cdot  P \left( \tilde{y} \right)+ \left(1-\epsilon  \right) \cdot P \left( y \right), \epsilon \sim U \left[0,1\right],
\end{split}
\label{equ:loss of Dp}
\end{equation}
where $P$ is the partition operation, $\lambda_{gp1}$ and $\lambda_{gp2}$ are the weighting parameters used to control the gradient penalty. Note that we apply both discriminators to the outputs at three different scales.

In addition to the adversarial loss, we also incorporate reconstruction loss $L_{rec}$, defined as follows: 
\begin{equation}
\begin{split}
\mathcal{L}_{rec}=\lambda _{l1}\left\| y-\tilde{y} \right\| _1+\lambda _{per}\left\| \psi _{vgg}\left( y \right) -\psi _{vgg}\left( \tilde{y} \right) \right\| _2,
\end{split}
\label{equ:loss of L1}
\end{equation}
where the $\psi_{vgg}$ represents a pre-trained VGG-19 network \cite{simonyan2014very}, from which we use feature maps from the 8$^{th}$, 15$^{th}$, and 22$^{nd}$ ReLU layers to compare shallow textures and deep features between real and generated images.

\subsection{Text Reconstruction Module}
We also incorporate a Text Reconstruction Module $T$ specifically designed to enhance our model's ability to accurately interpret characters on license plate images. $T$ merges the Decoder's intermediate feature $F_D$ with the fusion latent code  ${F_{2,3}}$ along the channel dimension. This combined feature then traverses a series of convolutional and linear layers, resulting in a vector that represents the recognized text. Concurrently, a pre-trained CRNN model extracts ground-truth text vectors from real images. We calculate the $L_1$ loss between the output vector from our Text Reconstruction Module T and the ground-truth text vector. The loss is defined as:
\begin{equation}
\begin{split}
\mathcal{L}_{text}=\left\| T \left( F_{2,3}, F_{D} \right)-\psi_{crnn}\left( y \right) \right\| _1,
\end{split}
\label{equ:loss of text}
\end{equation}
where $F_{D}$ is the feature maps obtained from the middle layers of Decoder, $\psi_{crnn}$ represents the pre-trained CRNN model.

\subsection{Fully Objective}
Our full objective is
\begin{equation}
\begin{split}
\mathcal{L}(E, D, F, D_g, D_p, T) &= \mathcal{L}_{rec} + \lambda_{g} \mathcal{L}_{D_g} + \lambda_{p} \mathcal{L}_{D_p} \\
&\quad + \lambda_{t} \mathcal{L}_{text},
\end{split}
\label{eq:loss}
\end{equation}
where $\lambda_{g}$, $\lambda_{p}$ and $\lambda_{t}$ control the relative importance of the different objectives. We aim to solve:

\begin{equation}
E^*, F^*, D^* = \arg\min_{\small E, F, D, T} \max_{D_g, D_p} \mathcal{L}(E, D, F, D_g, D_p, T).
\label{equ:loss_of_text}
\end{equation}

\section{Experiment}
\subsection{Experimental Setups}
\paragraph{Dataset setup.} The proposed \dataset dataset is partitioned into a training set with 9,288 image pairs, and a test set with 1,000 image pairs. The test set encompasses 500 pairs acquired under normal light conditions and another 500 pairs captured in low light conditions.

\paragraph{Evaluation metrics.} To evaluate the image quality of deblurred images, we adopt three evaluation metrics: PSNR, SSIM, and Perceptual Loss (PerL). The Perceptual Loss is specifically defined by comparing the generated images with the ground truth images at the output feature maps of sequential layers 8, 15, and 22 of a pre-trained VGG-19 model. To assess the recognisability of the generated license plate images, we calculate the Text Levenshtein Distance (TLD) \cite{levenshtein1966binary} between the detected text of the generated images and the real images.

\paragraph{Implementation details.}
The shape of multi-scale input images for \model are $(112, 224, 3)$, $(56, 112, 3)$, and $(28, 56, 3)$ respectively. Random rain adding and random cutout are utilized for data augmentation. The optimizer we use is Adam \cite{kingma2014adam}. The batch size is set to $7$. The initial learning rate is $10^{-4}$, and the linear weight decay is used after the $100{th}$ epoch. All experiments are conducted on a GeForce RTX 3090 GPU.

\subsection{Deblur Results}
To evaluate the deblur performance of our method, we compare \model with five SOTA methods: Pix2Pix \cite{isola2017image}, DeblurGAN v2 \cite{kupyn2019deblurgan}, MIMO-Unet \cite{cho2021rethinking}, MSSNet \cite{kim2022mssnet}, and LBAG \cite{li2023real} on \dataset. 

From the results presented in Table \ref{tab: total compare}, it can be observed that our \model outperforms all other models in both normal and low light conditions. In normal light conditions, our \model achieves a PerL of 3.31, PSNR of 29.95, and SSIM of 0.795, which are superior to the latest deblurring method LBAG. The performance gap becomes even more pronounced when dealing with low light images, with our model exhibiting improvements of 29.8\%, 4.5\%, and 3.7\% in PerL, PSNR, and SSIM, respectively, compared to LBAG and MSSNet.

Figure \ref{Fig:result} provides a visual comparison between sets of blurred and deblurred images under two light conditions. In the case of normal light, our \model effectively restores license plates afflicted with severe motion blur, accurately generating and reconstructing characters such as `D', `O' and `B', as well as numbers like `7', `1', and `L', which often pose challenges for other models, as shown in the $1st$, $3rd$ and $7th$ column of Figure \ref{Fig:result}. In low light scenarios, where license plates are barely visible to the human eye, our model excels in generating details that significantly surpass the performance of other models. This highlights the inadequacy of models designed for minor blurs in large scenes when applied to the deblurring of license plates, which are subject to more severe blurring.
\begin{figure}[t]
		\centering
		\includegraphics[width=1\linewidth]{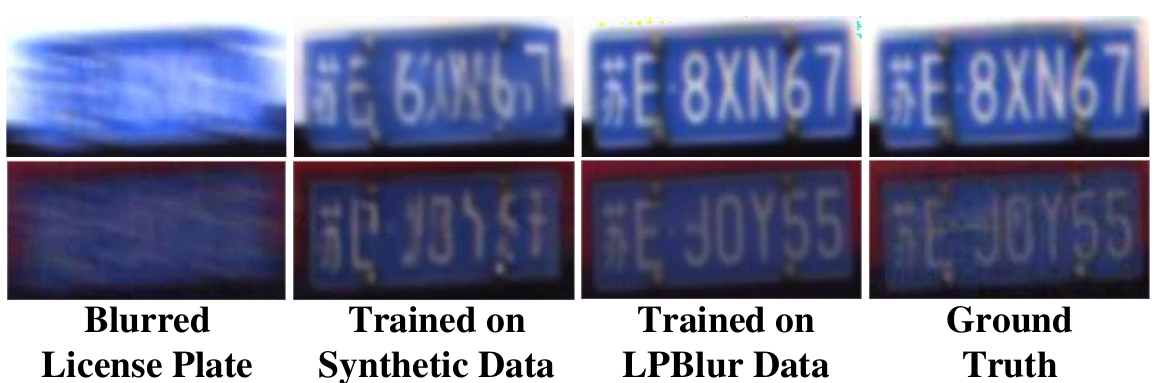} 
	\caption{Visual results comparison of \model trained on synthetic data and \dataset data.}
	\label{Fig:synthetic data}
\end{figure}
\begin{table}
\begin{tabular*}{\linewidth}{ccccc}
   \toprule
   Normal Light & PerL$\downarrow$ & PSNR$\uparrow$ & SSIM$\uparrow$ & TLD$\downarrow$ \\
   \midrule
   Synthetic Data  & 3.45 & 28.74 & 0.7061 & 1.68 \\
   \dataset Data & \textbf{3.31} & \textbf{29.95} & \textbf{0.7950} & \textbf{0.57} \\
   \toprule
   Low Light & PerL$\downarrow$ & PSNR$\uparrow$ & SSIM$\uparrow$ & TLD$\downarrow$ \\
   \midrule
   Synthetic Data  & 2.05 & 28.28 & 0.7933 & 2.65 \\
   \dataset Data & \textbf{1.01} & \textbf{30.96} & \textbf{0.9214} & \textbf{0.81} \\
   \bottomrule
\end{tabular*}
\caption{Quantitative comparison using synthetic and \dataset data in normal and low light scenarios, respectively.}
\label{tab: synthetic data}
\end{table}
\subsection{Text Recognition Results}
We further evaluate the plate text recognition accuracy based on those deblurred images. The CRNN \cite{shi2016end} is incorporated for the recognition of generated and sharp license plate characters. The $4th$ and $8th$ columns in Table \ref{tab: total compare} compare the TLD between the generated images and original sharp images. In the context of normal light conditions, \model exhibits comparable performance to LBAG in terms of TLD but surpasses all other models. Under low light conditions, \model is the only model with a TLD lower than 1. This implies that a pre-trained CRNN model, when employed to recognize deblurred license plate images produced by \model, will obtain an average error in less than one character per instance. Consequently, \model has the best performance overall, demonstrating robust capability in deblurring license plates across two scenarios.
\begin{table}

\tabcolsep=0.012\linewidth
\begin{tabular*}{\linewidth}{cccccccc}
   \toprule
   \multirow{3}*{Model No.} & \multicolumn{7}{c}{Normal Light} \\
   \cmidrule(lr){2-8} &$La.$ &$Te.$ &$PD$ & PerL$\downarrow$ & PSNR$\uparrow$ & SSIM$\uparrow$ & TLD$\downarrow$ \\
   \midrule
   1 & &\checkmark &\checkmark  & 3.49 & 29.68 & 0.7883 & 0.69 \\
   2 &\checkmark & &\checkmark  & 3.56 & 29.41 & 0.7797 & 0.72 \\
   3 &\checkmark &\checkmark &  & 3.41 & 29.73 & 0.7829 & 0.61 \\
   \model &\checkmark &\checkmark &\checkmark & \textbf{3.31} & \textbf{29.95} & \textbf{0.7950} & \textbf{0.57} \\
   \toprule
   \multirow{3}*{Model No.} & \multicolumn{7}{c}{Low Light} \\
   \cmidrule(lr){2-8} &$La.$ &$Te.$ &$PD$ & PerL$\downarrow$ & PSNR$\uparrow$ & SSIM$\uparrow$ & TLD$\downarrow$ \\
   \midrule
   1 & &\checkmark &\checkmark  & 1.26 & 30.05 & 0.9165 & 0.92 \\
   2 &\checkmark & &\checkmark  & 1.79 & 29.12 & 0.8861 & 1.45 \\
   3 &\checkmark &\checkmark &  & 1.38 & 29.93 & 0.9012 & 1.01 \\
   \model &\checkmark &\checkmark &\checkmark & \textbf{1.01} & \textbf{30.96} & \textbf{0.9214} & \textbf{0.81} \\
   \bottomrule
\end{tabular*}
\caption{Ablations of \model on \dataset. $La.$,$Te.$ and $PD$ denote the Latent Fusion Module, Text Reconstruction Module and Partition Discriminator Module, respectively.}
\label{tab: ablation}
\end{table}

\subsection{Ablation Study}
To evaluate the effectiveness of each proposed module, a series of ablation experiments are performed, which is shown in Table \ref{tab: ablation}. The omission of the Latent Fusion Module leads to a decline in global metrics, underscoring its effectiveness in fusing multi-scale features and improving the model's performance in restoring sharp images. Removing the Text Reconstruction Module results in a significant downturn in global metrics, particularly noticeable under low light conditions. This highlights the pivotal role of the Text Reconstruction Module in enabling the model to have a deeper understanding and restoration capability for license plates affected by severe pixel disruption. Similarly, the exclusion of the Partition Discriminator Module deteriorates global metrics and notably affects the SSIM metric to a greater extent. This confirms the module's contribution to enhancing the model focus on the details of each letter on the license plate.

\subsection{Necessity of \dataset}
We further demonstrate the importance of introducing a dataset consisting of real blurred images for the task of license plate deblurring. To assess this, we employ different blur kernels randomly to the sharp images in \dataset dataset and finally create a synthetic dataset. The result samples, illustrated in Figure \ref{Fig:synthetic data} and summarized in Table \ref{tab: synthetic data}, clearly indicate that \model trained on the synthetic dataset fails to eliminate real-world license plate blur effectively. This is evident both visually and in the metric evaluations, showcasing poorer performance compared to when trained on the \dataset dataset.

The aforementioned outcomes highlight a significant disparity between synthetic and real-world license plate blur, emphasizing that synthetic blurred image data cannot serve as a substitute for the \dataset dataset. Thus, the \dataset dataset proves to be more effective in training models for real-world license plate deblurring.

\section{Conclusion}
In this paper, we study the issue of motion license plates deblurring. We introduce the first large-scale license plate deblurring dataset for this research and address color bias and misalignment problems through appropriate data collection methods and post-processing. Furthermore, given that the degree of blur caused by vehicular motion substantially exceeds that induced by camera shake, we propose a model based on multi-scale input and output for license plate deblurring. This includes a latent fusion module, a supervision module for textual modality information, and a partition discriminator module. Experimental results indicate that our model performs favorably in comparison to current state-of-the-art deblurring algorithms. In the future, we intend to augment our dataset with license plates from a broader range of countries and regions to enhance its diversity. Regarding the model, we intend to incorporate modules that ensure the restoration capability for spatially complex characters, such as Chinese characters.

\appendix

\section*{Acknowledgments}
This work was jointly supported by the National Natural Science Foundation of China (62201474 and 62206180), Suzhou Science and Technology Development Planning Programme (Grant No.ZXL2023171) and XJTLU Research Development Funds (RDF-21-02-084, RDF-22-01-129, RDF-22-01-134, and RDF-23-01-053).

\section*{Ethical Statement}
To prevent the disclosure of personal privacy, all private information, including human faces and surrounding scenery, is removed from the images in the \dataset, and only the license plate number is retained. In addition, sensitive metadata in the image is removed, including GPS location, timestamp, etc.

\bibliographystyle{named}
\bibliography{ijcai24}

\end{document}